\documentclass[twoside]{article}
\usepackage[accepted]{aistats2012}
\usepackage{amsmath}
\usepackage[utf8]{inputenc}
\usepackage{cite}
\usepackage{comment}
\usepackage[colorinlistoftodos,prependcaption,textsize=tiny]{todonotes}
\usepackage{blindtext}
\usepackage[]{algorithm2e}
\usepackage{graphicx}
\usepackage{txfonts} 
\usepackage{enumitem}
\let\vec\mathbf
\let\mat\mathbf
\newcommand\given[1][]{\:#1\vert\:}

\newcommand{\card}[1]{\left\vert#1\right\vert}

\expandafter\def\expandafter\normalsize\expandafter{%
    \normalsize
    \setlength\abovedisplayskip{10pt}
    \setlength\belowdisplayskip{10pt}
    \setlength\abovedisplayshortskip{0pt}
    \setlength\belowdisplayshortskip{0pt}
}

\begin{document}

%

%

\twocolumn[

\aistatstitle{Probabilistic Trajectory Segmentation by Means of Hierarchical Dirichlet Process Switching Linear Dynamical Systems}

\aistatsauthor{Maximilian Sieb\footnotemark[1]   \And Matthias Schultheis\footnotemark[1] \And Sebastian Szelag\footnotemark[1] \And Rudolf Lioutikov \And Jan Peters}

\aistatsaddress{Technische Universität Darmstadt} ]
\footnotetext[1]{equal contribution}
\begin{abstract}
  

  Using movement primitive libraries is an effective means to enable robots to solve more complex tasks. In order to build these movement libraries, current algorithms require a prior segmentation of the demonstration trajectories. A promising approach is to model the trajectory as being generated by a set of Switching Linear Dynamical Systems and inferring a meaningful segmentation  by inspecting the transition points characterized by the switching dynamics. With respect to the learning, a nonparametric Bayesian approach is employed utilizing a Gibbs sampler. 

\end{abstract}
\section{Introduction}
In recent years, problems aimed to be solved by robots have become more and more complex. These problems, which include challenging tasks such as rescuing humans in crisis zones \cite{matsuno2004rescue} or playing table tennis \cite{muelling2010learning}, require the robot to act autonomously and infer the correct movements. In most of these complex scenarios, tasks cannot be precisely planned by hand, and thus, it has become worthwhile to let the robot learn the desired behavior from demonstration data.
One well-known concept for learning is \textit{imitation learning}, where a robot learns to perform tasks by imitating demonstrations provided by some teacher.
By learning the trajectory as a whole may enable the robot to reproduce it, but this approach does not generalize well and fails to perform tasks that require variability and variance.
To learn trajectories in a more versatile and generalizable manner, the trajectory can be divided into several subtasks which can be solved by means of elementary movements, referred to as movement primitives \cite{schaal2005learning}. The recently introduced ProbS algorithm \cite{lioutikov2015probabilistic} makes use of trajectories, which are for example collected from demonstrations, to build such a library.
However, this algorithm requires a meaningful segmentation initialization of the trajectories. Simple heuristic methods for providing initializations, such as segmentation at zero-velocity points, were stated to only be meaningful in a very restricted set of problems \cite{lioutikov2015probabilistic}.

\begin{figure}[h]
\centering
\includegraphics[width=\linewidth]{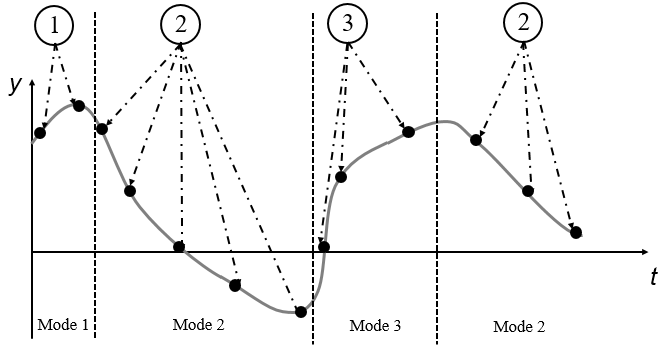}
\caption{\textbf{Trajectory segmentation} The algorithm will assign every time step to a specific mode while strongly enforcing self-transitions.}
\label{fig:traj-segmentation}
\end{figure}

The goal of our project is to provide a well-founded probabilistic approach for trajectory segmentation by means of Switching Linear Dynamical Systems (SLDS). The concept of SLDS is to model an observed time series as being generated by a set of different time-invariant dynamical systems. This method for trajectory segmentation can then be used as an adaptive way of providing an initial segmentation of the trajectory for the ProbS algorithm. In this work, we present a promising approach to apply the SLDS framework to the problem of partitioning trajectories by assuming the trajectory has been generated by such a generative model (see Figure \ref{fig:traj-segmentation}). The preliminary goal is to learn the parameters of the dynamical subsystems in the SLDS framework while simultaneously learning the specific points of time where the switching occurs, the so-called transition points.

While there exist many methods for learning SLDS models, most of them come with the downside of assuming a fixed number of dynamical modes to apply an Expectation Maximization (EM) algorithm. To circumvent this problem, we utilize a nonparametric Bayesian approach and make use of the concept of Hierarchical Dirichlet Processes (HDP) within the SLDS framework which was initially introduced by Fox\,et\,al.\ \cite{fox2009nonparametric, fox2011bayesian}. 
\\\\
In Sec. 3, we lay out the theoretical foundations relevant to the understanding of the algorithm. In Sec. 4, we explain how our setup infers and samples the desired quantities of SLDS model. In Sec. 5 we analyze the algorithm by means of a toy problem, where we use a randomly generated trajectory with switched dynamics.

\section{Related Work}

For learning the SLDS model in general applications, the following quantities are unknown and have to be learned: The dynamical parameters, which will be formally introduced in the following sections, the modes indicating the dynamical system for a time step, and the internal states which represent the noise-free trajectory data. These quantities are highly coupled, making the process of learning a challenging task. Therefore, in most of the related work, assumptions are made to make these problems more efficient to solve. One common simplification to the problem is to include prior knowledge about the number of modes. Even in this case, inference in SLDS can be shown to be formally intractable \cite{bar1993estimation}, and because of that, standard forward-backward algorithms cannot be applied as in the related Hidden Markov Model (HMM) learning problems. As the modes and the internal states can be seen as latent variables, it was suggested \cite{murphy1998switching, nemati2013learning} to use an Expectation Maximization (EM) approach for learning. This approach consists of two steps being iterated in an alternating manner: During the Expectation step, one firstly obtains the expected values of the latent variables using a modified Kalman smoother. Secondly, the model parameters have to be estimated 
during the maximization step by maximizing the expected complete data log-likelihood. For performing inference, which is required for computing the expectations, there are principally two methods available: Expectation Propagation (EP) \cite{minka2001family} and Generalized Pseudo Bayes (GPB) \cite{murphy1998learning, bar1993estimation}. With both approaches come certain limitations, e.\,g.\ requiring strict assumptions and the display of numerical instabilities. These, however, can be further relaxed by means of Expectation Correction (EC) \cite{barber2006expectation}. Oh et al.\ \cite{oh2005learning} extended the SLDS model further by introducing a parametric SLDS, introducing global parameters representing variations in motions of honey bees.
These approaches were applied to various applications such as dancing honey bees, human motion recognition \cite{pavlovic2000learning, ren2005data} and even number 
recognition in voice recordings \cite{mesot2007switching}, where the number of switching states can be fixed.
For the segmentation of general robot trajectories, however, it is disadvantageous to fix the number of modes. Generally, it is unknown how many movement primitives a trajectory consists of and how complex they are, and thus, the underlying number of modes varies in every case. Therefore, in this application scenario, more general methods need to be employed. Huang\,et\,al.\ \cite{huang2004identification} approach this problem by embedding the input and output data in a higher-dimensional space and they identify the points in time where switching occurs by segmenting the data into distinct subspaces. This approach, however, assumes deterministic dynamics which is usually not suitable in practice. Fox\,et\,al.\  \cite{fox2009nonparametric, fox2011bayesian} opted for a nonparametric approach using Hierarchical Dirichlet Processes (HDPs) in order to learn the number of modes without prior specification of the overall number of modes within the system. In their work, the modes, internal states and model parameters are alternately sampled conditioned on all other quantities. The main drawback of the method is that it is computationally expensive and requires a large amount of tuning of the parameters.

\section{Foundations}

In this section, we first formally introduce the SLDS model. Also, the theoretical foundations of the Dirichlet Process and the extension to the Hierarchical Dirichlet Process are explained. We also address the problem of high-frequent switching of the modes that comes with an unaltered HDP-approach. The section is then concluded by explaining a variation of a Kalman Filter that is used to smooth the input data.

We decided to investigate the HDP-based method by Fox\,et\,al.\ \cite{fox2009nonparametric, fox2011bayesian} for our problem to segment robot trajectories. The main reason for this is that we want to alleviate the previously mentioned restriction of having to fix the number of modes beforehand. 
Fixing the number of modes would cause trajectories of different complexities to be represented with the same number of dynamical systems mostly leading either to an underestimation or an overestimated of the actual number of underlying modes. On the one hand, assuming too few dynamical systems would lead to a bad approximation of the trajectory and would need further subdivision. On the other hand, assuming a too granular segmentation,leading to a smaller approximation error at the expense of introducing a larger model complexity, would neither yield a meaningful segmentation because it would exhibit high-frequency switching behavior over short time periods.
Furthermore, the HDP-based SLDS approach is described to be more robust to noise which is an important property when dealing with inherently noisy observations.

\subsection{SLDS}
\label{sec:slds}

\begin{figure}[h]
\centering
\includegraphics[width=.7\linewidth]{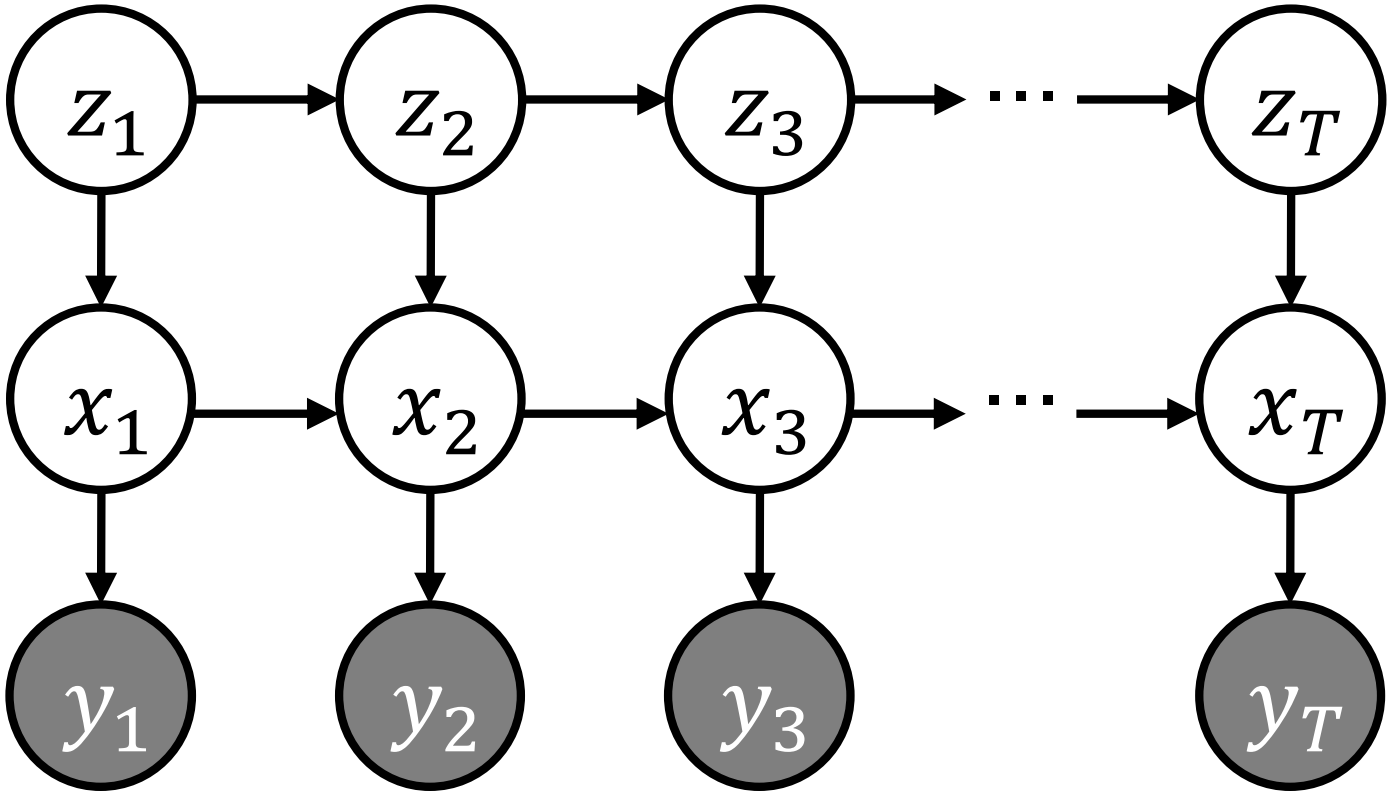}
\caption{\textbf{SLDS graphical model} In an SLDS the observations $\vec{y}_{1:T}$ are modelled to be generated by a dynamical system with states $\vec{x}$ and switching dynamics $\mat{A}^{(z_t)}$.}
\label{fig:slds-model}
\end{figure}

Switching Linear Dynamical Systems (SLDS) model time-discrete observations $\vec{y}_t$ as being generated by a linear dynamical system whose dynamics switch over time (see Figure \ref{fig:slds-model}). The observations $\mathbf{y}_t$ are generated by a linear transformation of internal states $\mathbf{x}_t$ and additional Gaussian noise $\mathbf{w}_t \sim \mathcal{N}(0, \mat{R})$: $\mathbf{y}_t = \mathbf{C}\mathbf{x}_t + \mathbf{w}_t$. The state dynamics are given by $\mathbf{x}_t = \mathbf{A}^{(z_t)} \mathbf{x}_{t-1} + \mathbf{e}^{(z_t)}_t$ with the state transition noise $\mat{e}_t^{(z_t)}\sim \mathcal{N}(0,\mat{\Sigma}^{(z_t)})$. $z_t$ designates the mode describing which dynamical matrix $\mathbf{A}^{(z_t)}$ is applied at this time step. We assume for each mode a distribution for generating the mode of the following time step: $z_t \sim \pi_{z_{t-1}}.$

\subsection{Dirichlet Process}
\label{sec:dp}

In order to model the previously introduced distributions $\pi$ for the different dynamical modes, we use a Dirichlet Process (DP). The DP can be described as a measure on a measure \cite{teh2004sharing}, and it samples discrete probability measures from an underlying base measure $G_0$. An explicit draw $G\sim DP(\gamma,G_0)$ from this DP can be expressed as $G=\sum^{\infty}_{k=1}\beta_k\delta_{\theta_k}$, where $\theta_k$ is given by $\theta_k \sim G_0$ and $\beta_k$ is the corresponding probability weight of the associated $\theta_k$ \cite{sethuraman1994constructive}. The discrete probability measure for $\beta$, denoted by $\mat{\beta} \sim GEM(\gamma)$, is defined as
\begin{align*}
\hspace{.5cm}v_k&\sim Beta(1,\gamma) &k=1,2,...\\
\beta_k&=v_k\prod_{l=1}^{k-1}(1-v_l) &k=1,2,...
\end{align*}
A naive approach would be to sample a single probability distribution $G$ from the DP governing the distribution of the dynamical parameters for the entire time series. The mode-specific dynamical parameters would then be sampled each time step from $G$:
\begin{align*}
G&\sim DP(\gamma,G_0)\\
\theta^{\prime}_t\given G&\sim G\\
x_t\given\theta^{\prime}_t&\sim F(\theta^{\prime}_t)
\end{align*}
As shown in \cite{teh2004sharing}, the DP draw $G$ can be expressed through stick-breaking construction $G=\sum^{\infty}_{k=1}\beta_k\delta_{\theta_k}$, and therefore, an equivalent representation of the DP is given by
\begin{align*}
\mat{\beta} &\sim GEM(\gamma)\\
\theta_k &\sim G_0\\
z_t &\sim\mat{\beta}\\
x_t &\sim F(\theta_{z_t})
\end{align*}
where $F(\theta_i)$ is the parametrization of our model, which is given by $x_t \sim \mathcal{N}(\mat{A}^{(z_t)}x_{t-1},\mat{\Sigma}^{(z_t)})$. The base measure $G_0$ is the modelled distribution for the dynamical parameters and is discussed in greater detail in later chapters. 
Comparing the two representations outlined above, it becomes apparent that the observed dynamical parameters $\theta^{\prime}_t$ for each time step assume a specific $\theta_k$ with probability $\beta_k$. Because of this nonparametric setup, the number of modes does not have to be fixed before hand.

Using the DP as outlined above to represent our mode transition probabilities, we face the problem of having the same probability distribution for every mode. We can enhance our model to draw a new probability measure $G_t\sim DP(\gamma,G_0)$ in each time step. However, by assuming a continuous base measure $G_0$, the probability distributions for each time step will now have no dynamical parameters in common, i.\,e.\ $\Theta_i \cap \Theta_{j:\;j\neq i} =\emptyset$. Hence, every time step will have different dynamical parameters which defeats the purpose of our model to find transition points of dynamical modes. This problem can be overcome by introducing a discrete base measure $G_0$ resulting in a shared parameter space $\Theta$ for each time step as discussed in the next chapter.

\subsection{Hierarchical Dirichlet Process}
So far, we could sample an entire mode probability distribution from our Dirichlet prior for every time step. The major drawback of this approach, as discussed previously, is that these draws will not share any similarities, since  the entire parameter space $\Theta_k$ for each probability measure drawn from a DP with a continuous base distribution $H$ will satisfy $\Theta_i \cap \Theta_{j:\;j\neq i} =\emptyset$. Therefore, we extend our model to a Hierarchical Dirichlet Process, allowing for shared similarities across multiple modes.
To do so, we model our base distribution $G_0$ as a probability measure drawn from another Dirichlet process $DP(\gamma,H)$. Our intermediate probability draws $G_j$ are then again draws from the intermediate distribution $G_0$, given by $G_j\sim DP(\alpha, G_0)$. Also, we do not use a different distribution for each time step, but rather assign an indicator variable $z_t$ to every time step that represents the current mode of the system. Thus, every mode has its specific mode transition probability distribution $G_j$ along with its associated dynamical parameters $\theta_j$. The modified model is now given by
\begin{align*}
G_0&\sim DP(\gamma,H)\\
G_j&\sim DP(\alpha,G_0)\\
\theta^{\prime}_t&\sim G_{j:\;\theta^{\prime}_{t-1}=\theta_j}\\
x_t&\sim F(\theta^{\prime}_t).
\end{align*}
One can integrate out the intermediate distribution $G_0$, as shown in \cite{teh2004sharing} and use the stick-breaking representations $G_j=\sum^{\infty}_{k=1}\pi_{j k}\delta_{\theta_k}$ and $G_0=\sum^{\infty}_{k=1}\beta_k\delta_{\theta_k}$ to get an equivalent representation of the HDP given by
\begin{align*}
\mat\beta&\sim GEM(\gamma)\\
\mat\pi_j&\sim DP(\alpha,\beta)\\
z_t \given z_{t-1}&\sim \mat\pi_{z_{t-1}}\\
\theta_k&\sim H\\
x_t&\sim F(\theta_{z_t})
\end{align*}
with $\theta^{\prime}_t=\theta_{z_t}$. Here, $\theta^{\prime}_t$ describes the dynamical parameters for the current time step and current mode, respectively, whereas $\theta_k$ describes the entire underlying family of dynamical modes available to be drawn from. It can be easily seen that $\Theta_i \cap \Theta_{j:\;j\neq i} \neq\emptyset$ because all $G_j$ now share the same support points $\theta_k$ from the discrete parental distribution $G_0$.\\\\
With regards to practical implementations, the infinite mixture model depicted by the HDP has to be approximated. Suppose we have a mixture model with $L$ mixture components and mixing weights $\mat{\beta}=(\beta_1,...,\beta_L)$.  This finite mixture model has the Dirichlet distribution $Dir(\gamma/L,...,\gamma/L)$ as a conjugate prior. As laid out in \cite{teh2004sharing}, this Dirichlet distribution is an equivalent representation of a DP if $L\rightarrow\infty$. Thus, the mixture weights $\beta$ of the finite mixture model can adequately represent a DP if $L$ is chosen sufficiently big. 
Using this so-called \textit{weak limit approximation}, the probability weights of our HDP can now be expressed as
\begin{align*}
\mat{\beta}&\sim Dir(\gamma/L,...,\gamma/L)\\
\mat{\pi}_j&\sim Dir(\alpha \beta_1,...,\alpha\beta_L)
\end{align*}This representation is useful as it allows to draw weights from a finite probability distribution while also providing an efficient means of computing the posterior probability for later inference.

\subsection{Sticky Extension for HDP}

Introducing the HDP prior enables our model to have recurring dynamical modes across multiple times steps. Furthermore, as shown in \cite{fox2011bayesian}, the expectation of the transition probabilities $\mat{\pi}_j$ is identical for all mode-specific distributions:
\begin{align*}
\mathbb{E}[\pi_{j k}\given\mat{\beta}]=\beta_k
\end{align*}
The main problem now is that the current setup displays fast switching between all time steps. This comes to no surprise since the mode-specific transition distributions currently do not favor self-transitions. Keeping our goal in mind, namely finding a segmentation for a demonstration task, we have to account for high-probability self-transitions; otherwise, a meaningful segmentation is almost impossible to be obtained. Therefore, in a similar manner to \cite{fox2009nonparametric}, we modify the mode-specific transition distributions $\mat{\pi}_j$ with a \textit{stickiness}-parameter $\kappa$ as follows:
\begin{align*}
\mat{\pi}_j\sim DP\left(\alpha+\kappa,\frac{\alpha\mat{\beta}+\kappa\delta_j}{\alpha+\kappa}\right)
\end{align*}
The parameter $\kappa>0$ represents the probability for a self-transition $z_t = z_{t-1}$. If $\kappa=0$, the original non-sticky HDP is recovered.
Incorporating $\kappa$ into the weak limit approximation yields
\begin{align*}
\mat{\pi}_j\sim Dir(\alpha \beta_1,...,\alpha \beta_j+\kappa,...,\alpha\beta_L).
\end{align*}

\subsection{Kalman Filtering}

The observed trajectory data is inherently noisy.
Therefore, we utilize a Kalman filter to smooth the observed data to estimate the internal states $x_t$ of the SLDS model. The Kalman filter provides a recursive algorithm for estimating the underlying state of a linear-Gaussian state space model in SLDS given a set of observations and fixed model parameters\cite{Ebfox2009slds}. It can be used to predict quantities making use of both the noisy observations and the model knowledge.
Applying the Kalman filter to the observations or, more specifically, the recorded trajectories of the SLDS, we preliminarily assume that the dependency between $y_t$ and $x_t$ is a linear transformation with added Gaussian noise, that is $y_t = Cx_t + w_t$ with $w_t\sim \mathcal{N}(0,R)$ (see section \ref{sec:slds}), where we set $\mat{C} = \mat{I}$ to model the observations simply as a noisy representation of the hidden states. 
We employ a combination of a forward and backward Kalman filter, which uses the previously sampled states together with the sampled dynamical matrices $A^{(z_t)}$ as well as the observations $y_t$, to predict the states $x_{1:T}$ for the respective current iteration. This variant of the Kalman filter is also known as the Rauch-Tung-Striebel Kalman smoother \cite{Ebfox2009slds}.

\section{Inference and Learning in SLDS}
\label{sec:learning-slds}

Inference in SLDS can be shown to be formally intractable \cite{bar1993estimation}, and therefore, we have to resort to sampling methods. We employ Gibbs sampling to sample the variables of interest, namely the state sequence $x_{1:T}$, the mode sequence $z_{1:T}$ and the dynamical parameters described by the parametric family $\theta_{1:K}$, where $K$ is the overall number of different dynamical modes estimated to compose our system. Note that $K \leq L$, where $K$ is the attained number of modes over the entire time series while $L$ merely represents the truncated limit of the maximum number of modes that can possibly occur. Additionally, we update and sample from the posterior of the underlying transition probability distribution $\beta$, the posterior of the mode-specific transition probability distributions $\pi_k$ and the posterior of the hyperparameters $\gamma$, $\alpha+\kappa$ and $\rho$. In general, the process can be described by the following steps:

\begin{enumerate}[itemsep=-2pt,start=0]
\item Initialize all values\\
\textbf{Repeat:}
\item Sequential sampling of $z_{1:T}$
\item Block sampling of $x_{1:T}$
\item Block sampling of $z_{1:T}$
\item Sampling of $\beta$ and all $\pi_k$
\item Sampling of hyperparameters $\gamma$, $\alpha$, $\kappa$, and $\rho$
\item Sampling of dynamical parameters $\mat{A}^{(k)}$ and $\mat{\Sigma}^{(k)}$ 
\end{enumerate}

\subsection{Gibbs Sampling}

In order to infer the transition probabilities $\pi$ along with the dynamical parameters $\theta$ and the state sequence $x_{1:T}$, we are sampling from their posterior distributions given the respective other quantities and incorporating a prior. Apart from the fact that deriving the MAP solution is analytically intractable, sampling from the posterior allows us to determine the expected value rather than a single best estimate which is a more appropriate measure for our application. For sampling we resort to Gibbs sampling.

Gibbs sampling allows to sample from the high-dimensional joint distribution by iteratively sampling from the conditional distributions of the quantities individually.
A more detailed introduction to Gibbs samplers as well as implementation details can be found in \cite{robert2004monte}.

\subsection{Block Sampling of Mode Sequence $z_{1:T}$}

As shown in \cite{fox2011bayesian}, the probability distribution of $z_t$, conditioned on the states $x_{1:T}$ and the set of dynamical parameters $\theta$, is given by \begin{center}
$p(z_t\;|\;z_{t-1},y_{1:T},\pi,\theta)\propto p(z_t\;|\;\pi_{z_{t-1}})p(x_t\;|\;\theta_{z_t})m_{t+1,t}(z_t)$.
\end{center}
The Marcovian structure of our model makes $z_t$ dependent only on the previous mode value $z_{t-1}$ through the transition probability $\pi_{z_{t-1}}$.
Recalling that $p(z_t\given\pi_{z_{t-1}}) = \pi_{z_{t-1}}$ and $p(x_t\given\theta_{z_t})=\mathcal{N}(x_t\given\mat{A}^{(z_t)}x_{t-1},\;\boldsymbol{\Sigma^{(z_t)}})$
, the entire joint auxiliary sampler for $z_{1:T}$ can be described as
\begin{align*}
z_t\sim\sum_{k=1}^L\pi_{z_{t-1}}(k)\mathcal{N}(x_t\;|\;\mat{A}^{(z_t)}x_{t-1},\;\boldsymbol{\Sigma^{(z_t)}})m_{t+1,t}(k)\delta(z_t,k)
\label{eq:2}
\end{align*}
where the mode sequence $z_{1:T}$ can be computed recursively starting at $z_0$.
The backward messages $m_{t+1,t}(z_{t-1})$ can also be recursively computed with
\begin{align*}
m_{t,t-1} \propto
\begin{cases}
    \sum_{z_t} p(z_t\;|\;\pi_{z_{t-1}})p(y_t\;|\;\theta_{z_t})m_{t+1,t}(z_t), & \text{if } t \leq T\\
    1 & \text{if } t = T+1\\
  \end{cases}
\end{align*}
Note that sampling each $z_t$ recursively is equivalent to joint sampling $z_{1:T}$ due to the Marcovian structure of our model.
By using a Dirichlet Process approximation for our model and taking advantage of its clustering property, only a small fraction of the maximum number of modes $L$ will be attained.

\subsection{Block Sampling of State Sequence $x_{1:T}$}
Conditioned on the mode sequence $z_{1:T}$ and the set of dynamical parameters $\theta$, the entire system degenerates into a simple linear dynamical system with switching dynamical parameters. As seen in \cite{fox2011bayesian}, the state sequence $x_{1:T}$ can be jointly sampled utilizing a  Kalman filter based approach. The probability of sampling $x_t$  conditioned on the prior state $x_{t-1}$ is given by
\begin{align*}
    p(x_t|x_{t-1},y_{1:T},z_{1:T},\theta)\propto p(x_t|x_{t-1},A^{(z_t)},\Sigma^{(z_t)})p(y_t|R)m_{t+1,t}(x_t)
\end{align*}
Recalling the SLDS setup of our system, this probability can be equivalently expressed as
\begin{align*}
    &p(x_t|x_{t-1},y_{1:T},z_{1:T},\theta)\propto\\ 
    &\qquad \mathcal{N}(x_t;A^{(z_t)}x_{t-1},\Sigma^{(z_t)})\mathcal{N}(y_t;Cx_t,R)m_{t+1,t}(x_t)
\end{align*}
The backward messages are given by 
\begin{align*}
    m_ {t+1,t}(x_t)\propto \mathcal{N}^{-1}(x_{t-1};\vartheta_{t,t-1},\Lambda_{t,t-1}),
\end{align*}
being initialized with $m_{T+1,T}\sim \mathcal{N}^{-1}(x_T;0,0)$.  Here, the so-called information parameters $\vartheta$ and $\Lambda$ can be recursively calculated prior to resampling $x_{1:T}$ since they only depend on the given observation sequence $y_{1:T}$ and the conditioned values for $\theta$ and $z_{1:T}$; refer to \cite{Ebfox2009slds} for implementation details.

Since the entire probability distribution is a simple product of Gaussian distributions, it can be further simplified using to yield a single Gaussian distribution for the block sampler $x_{1:T}$:
\begin{align*}
    x_t&\sim\mathcal{N}(x_t;\hat{\mu},\hat{\Sigma})\\
    \hat{\mu}&=(\Sigma^{(-z_t)}+\Lambda_{t|t}^b)^{-1}(\Sigma^{-(z_t)}A^{(z_t)}x_{t-1}+\theta_{t|t})\\
    \hat{\Sigma}&=(\Sigma^{-(z_t)}+\Lambda_{t|t}^b)^{-1}
\end{align*}
As can be seen, the entire state sequence $x_{1:T}$ can be recursively sampled starting at $x_0$. $\Lambda^f_{t|t}, \vartheta^b_{t|t}$ are generated from the backward filter. They contain the  marginalized observations  $x_{t+1:T}$. Due to the Markov property of our system, recursively sampling the state sequence is equivalent to jointly sampling the entire state sequence.

\subsection{Sequential Sampling of Mode Sequence $z_{1:T}$}

In order to improve the mixing rate of the Gibbs sampler, it has been shown to be fruitful \cite{Ebfox2009slds} to interleave an sequential sampling of $z_{1:T}$ from the distribution $p(z_t\;|\;z_{\setminus t},y_{1:T},\pi,\theta)$ prior to block sampling $x_{1:T}$. The implementation follows in principle the block sampling of $x_{1:T}$ by applying a forward and backward Kalman filter. By conditioning on all other modes except for the mode of the current time step $z_t$, and marginalizing over $x_{1:t-2}$, we obtain the distribution \begin{align*}
    p(x_{t-1}\;|\;y_{1:t-1},z_{1:t-1},\theta).
\end{align*}
Also, by marginalizing over $x_{t+1:T}$ we get 
\begin{align*}
    p(y_ {t+1:T}\;|\;x_t,z_{t+1:T},\theta).
\end{align*}
In principle, these distributions are equivalent to the backward and forward messages for the current time step. We combine these messages with the local likelihood of the current observation $p(y_t\;|\;x_t)$ and the transition probability $p(x_t\;|\;x_{t-1},\theta,z_t=k)$. Finally, we obtain the desired distribution by marginalizing over $x_t$ and $x_{t-1}$ and a subsequent multiplication with the probabilities of transitioning to $z_t=k$ from $z_{t-1}$ (given by $\pi_{z_{t-1}}(k)$) and from $z_t=k$ to $z_{t+1}$ (given by $\pi_k(z_{t+1})$).
The analytical expression for sampling $p(z_t=k\;|\;z_{\setminus t},y_{1:T},\pi,\theta)$ is then given by
\begin{align*}
    z_t\sim \sum_{k=1}^L\pi_{z_{t-1}}(k)\pi_k(z_{t+1})f_k(y_{1:T})\delta(z_t,k)
\end{align*}
with
\begin{align*}
&f_k(y_{1:T})=
\frac{|\Lambda_t^{(k)}|^{1/2}|}{\Lambda_t^{(k)}+\Lambda_{t|t}^b|^{1/2}}\\
&exp(-\frac{1}{2}\vartheta_t^{(k)^T}\Lambda_t^{-(k)}+\frac{1}{2}(\vartheta_t^{(k)}+\vartheta_{t|t}^b)^T(\Lambda_t^{(k)}+\Lambda_{t|t}^b)^{-1}(\vartheta_t^{(k)}+\vartheta_{t|t}^b))
\end{align*}
The information matrices $\vartheta$ and $\Lambda$ are obtained by running a forward and backward Kalman filter as mentioned above. \cite{Ebfox2009slds} shows explicit derivations for how to obtain these matrices.

\subsection{Sampling Dynamical Parameters}

Conditioned on the current mode sequence $z_{1:T}$ and state sequence $\vec{x}_{1:T}$, we can sample appropriate dynamic parameters $\mat{A}^{(k)}$ and $\mat{\Sigma}^{(k)}$ under incorporation of a prior.
To analyze the posterior distribution of the dynamic parameters, it is suitable to write the dynamic equation in the form
$\vec{x}_t = \mat{A}^{(k)} \ \vec{x}_{t-1} + \vec{e}_t$. Equivalently, we can state the equation in a linear regression form if we form for every $k$ a matrix $\mat{X}^{(k)}$ with $N_k$ columns consisting of the state vectors $\vec{x}_t$ for that $z_t = k$. In the same way, we construct a matrix $\mat{\bar{X}}^{(k)}$ consisting of the state vectors of the respective previous time steps, i.\,e. of the state vectors $\vec{x}_t$ for which $z_{t+1} = k$. Subsequently, we can formalize our problem as $$\mat{X} = \mat{A}^{(k)} \ \mat{\bar{X}}^{(k)} + \mat{E}^{(k)}$$ with $\mat{E}^{(k)}$ being a matrix consisting of Gaussian noise.

For inferring the dynamical parameters, we set a matrix-normal prior on $\mat{A}^{(k)}$ and a inverse-Wishart prior on the covariance matrices $\mat{\Sigma}^{(k)}$. This is a common choice in multivariate linear regression problems as the priors are conjugate in this case.

The matrix normal distribution is characterized by the probability density function $$ p(\mat{X} \given \mat{M}, \mat{U}, \mat{V}) = \frac{\exp\left( -\frac{1}{2} \mathrm{tr}\{\mat{V}^{-1} (\mat{X} - \mat{M}) ^T \mat{U}^{-1} (\mat{X} - \mat{M})\}\right)}{(2\pi)^{np/2} \card{\mat{V}}^{n/2} \card{\mat{U}}^{p/2} }$$ with mean matrix $\mat{M} \in \varmathbb{R}^{n \times p}$ and scale matrices $\mat{U} \in \varmathbb{R}^{n \times n}$, $\mat{V} \in \varmathbb{R}^{p \times p}$.

To bring the posterior distribution of the dynamical parameters in a form suitable for incorporating priors, we decompose it in the following way: $$ p\left(\mat{A}^{(k)}, \mat{\Sigma}^{(k)} \given \mat{D}^{(k)}\right) = p\left(\mat{A}^{(k)} \given \mat{\Sigma}^{(k)}, \mat{D}^{(k)}\right) \, p\left(\mat{\Sigma}^{(k)} \given \mat{D}^{(k)}\right)$$
The quantity $p\left(\mat{A}^{(k)} \given \mat{\Sigma}^{(k)}, \mat{D}^{(k)}\right)$ can be derived \cite{fox2011bayesian} in closed form by applying Bayes' formula using a prior and data likelihood which are both matrix-normal distributed: 
\begin{equation}
p\left(\mat{A}^{(k)} \given \mat{\Sigma}^{(k)}, \mat{D}^{(k)}\right) = \mathcal{MN} \left(\mat{A}^{(k)}; \mat{S}^{(k)}_{x\bar{x}} \left(\mat{S}^{(k)}_{\bar{x}\bar{x}}\right)^{-1}, \mat{\Sigma}^{(k)}, \mat{S}^{(k)}_{\bar{x}\bar{x}}\right)
\label{eq:dynsamp-a}
\end{equation}
with $\mat{S}^{(k)}_{\bar{x}\bar{x}} = \mat{\bar{X}}^{(k)}\mat{\bar{X}}^{(k)^T} + \mat{K}$ and $ \mat{S}^{(k)}_{x\bar{x}} = \mat{X}^{(k)} \mat{\bar{X}}^{(k)^T}  + \mat{M}\mat{K}$. $\mat{M}$ and $\mat{K}$ are parameter matrices determining the matrix-normal portion of the prior.


To obtain $\mat{\Sigma}^{(k)}$, we combine the matrix-normal distributed likelihood $p\left(\mat{D}^{(k)} \given \mat{\Sigma}^{(k)}\right)$ with a conjugate inverse-Wishart prior $\mathrm{IW}\left(n_0, \mat{S}_0\right)$. $\mat{S}_0$ is the scaling matrix and $n_0$  represents the degrees of freedom of the inverse-Wishart prior. The posterior is derived \cite{fox2011bayesian} to be 
\begin{equation} p\left(\mat{\Sigma}^{(k)} \given \mat{D}^{(k)}\right) = \mathrm{IW}\left( N_k + n_0,\ \mat{S}^{(k)}_{x\given\bar{x}} + \mat{S}_0 \right)
\label{eq:dynsamp-sigma}
\end{equation}

with $\mat{S}^{(k)}_{x\given\bar{x}} = \mat{S}^{(k)}_{xx} - \mat{S}^{(k)}_{x\bar{x}}\, {\mat{S}^{(k)}_{\bar{x}\bar{x}}}^{-1}\, {\mat{S}^{(k)}_{x\bar{x}}}^T$, where $\mat{S}^{(k)}_{xx} = \mat{X}^{(k)} {\mat{X}^{(k)}}^T + \mat{M}\mat{K}\mat{M}^T$ and $\mat{\bar{X}}^{(k)}\mat{\bar{X}}^{(k)^T}$, $\mat{X}^{(k)}\mat{\bar{X}}^{(k)^T}$ defined as above.

For sampling the dynamical parameters given $z_{1:T}$ and $\vec{x}_{1:T}$, we first sample $\mat{\Sigma}^{(k)}$ from the conditional distribution (\ref{eq:dynsamp-sigma}). Subsequently, we sample $\mat{A}^{(k)}$ from (\ref{eq:dynsamp-a}) conditioned on $\mat{\Sigma}$.

\subsection{Sampling of State Transition Distributions $\pi_k$ and Global Transition Distribution $\beta$}
Lastly, the mode transition probabilities are re-sampled using their respective posterior distributions. The global transition distribution $\beta$ is re-sampled from the posterior that is given by
\begin{center}
$\beta\sim Dir(\gamma/L+\bar{m}_{\cdot1},...,\gamma/L+\bar{m}_{\cdot L})$\\
\end{center}
The values for $\bar{m}_{jk}$ represent the number of transitions from mode $j$ to mode $k$ that were caused by a transition from a so-called \textit{informative} table also taking into account self-transition overrides. To fully understand this terminology, an introduction to the Chinese Restaurant Franchise Process is required. However, this would extend the scope of this paper and can therefore be studied in more detail in \cite{Ebfox2009slds} or \cite{teh2004sharing}.
The state transition distribution for each mode is re-sampled from the posterior distribution given by
\begin{center}
$\pi_k\sim Dir(\alpha\beta_1+n_{k1},...,\alpha\beta_k+\kappa+n_{kk},...,\alpha\beta_L+n_{kL})$
\end{center}
Here, $n_{jk}$ represents the number of transitions from mode $j$ to mode $k$ over the entire time series.



\section{Experiments}
We have implemented the Gibbs sampler for estimating the modes, internal states and dynamics as described in section \ref{sec:learning-slds} and evaluated its components in a toy setup. 

\begin{figure}
\centering
\includegraphics[width=\linewidth]{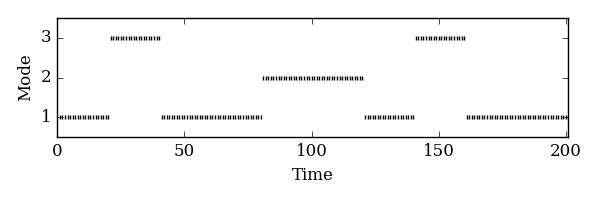}
\caption{\textbf{True mode sequence} The true mode sequence which was used to generate the trajectory.}
\label{fig:real-modes}
\end{figure}

The data for the toy problem is generated assuming a 2-dimensional state space and using the three randomly generated dynamical transition matrices
\begin{equation*}
A_1=\begin{bmatrix} 0 & 1\\ 0.7 & 0.36 \end{bmatrix},\ A_2=\begin{bmatrix} 0 & 1\\ 0.4 & 0.56 \end{bmatrix},\
A_3=\begin{bmatrix} 0.5 & 0.5\\ 0.32 & 0.67 \end{bmatrix}
\end{equation*}
as well as the state transition noise matrices 
\begin{equation*}
\Sigma_1=\Sigma_2=\Sigma_3 = \begin{bmatrix} 10^{-5} & 0\\ 0 & 10^{-5} \end{bmatrix}.
\end{equation*} This setup could be seen as observing two joint angles, for example, while trying to infer likely switching points. The state sequence was initialized with $x_0=1$ and generated following the equations of the Switching Linear Dynamical System (see section \ref{sec:slds}) with a pre-defined mode setup (see Figure \ref{fig:real-traj} and \ref{fig:real-modes}).

The initial SLDS parameters are sampled from the according priors:
\begin{align*}
    (\alpha + \kappa) \sim Gamma(a,b)\\
    \rho \sim Beta(c,d)\\
    \gamma \sim Gamma(e,f)
\end{align*}
$\alpha$ and $\kappa$ can be calculated from the deterministic relationships
\begin{align*}
    \alpha = (1- \rho)(\alpha + \kappa)\\
    \kappa = \rho(\alpha + \kappa).
\end{align*}
The hyperparameters of the sampler are set to $a = 10$, $b=1$, $c=20$, $d=2$, $e=10$, and $f=1$. These values allow for sufficiently spread out probability distributions $\pi_k$ ($a/b \gg 1$) and $\beta$ ($e/f\gg1$), and also account for a rather heavy sticky component ($c/d \gg 1$). 
The sampler itself is initialized with the previously defined priors. The initial $\beta$ and $\pi_k$ distributions are generated with a uniform Dirichlet prior with the given hyperparameters. The mode sequence is initialized according to the corresponding $\pi_k$ sample indicated by the previously sampled mode assignment $z_t$. The dynamical matrices are initialized by sampling from the according prior as described in section 4.5.

\begin{figure}
\centering
\includegraphics[width=\linewidth]{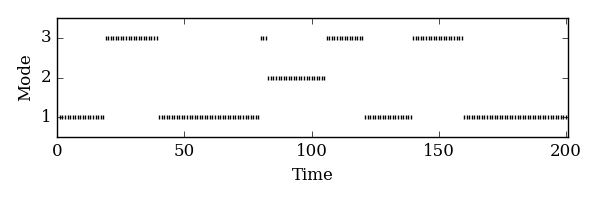}
\caption{\textbf{Sampled mode sequence} The sampled mode sequence provided by the algorithm. Most of the modes are consistent with the true ones.}
\label{fig:sampled-modes}
\end{figure}

\begin{figure}[h]
\centering
\includegraphics[width=\linewidth]{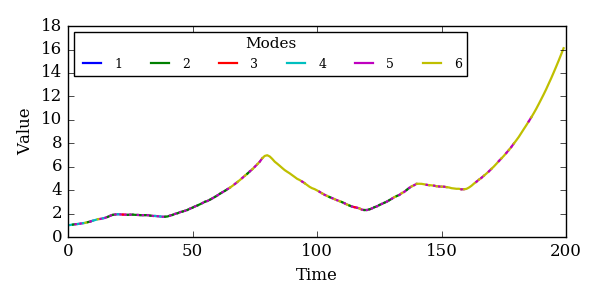}
\caption{\textbf{Non-sticky sampled mode sequence} The sampled mode sequence using the algorithm without sticky mode transitions ($\kappa = 0$). Note that only the first dimension of the state sequence is shown.}
\label{fig:sampled-modes-nonsticky}
\end{figure}

The system was run with the given setup for 105 iterations and the best sample of the last five iterations was taken. The resulting segmentation is shown in Figure \ref{fig:real-traj} and the underlying mode sequence in Figure \ref{fig:sampled-modes}. As can be seen, the sampler is able to match almost every mode correctly.
\begin{figure}
\centering
\includegraphics[width=\linewidth]{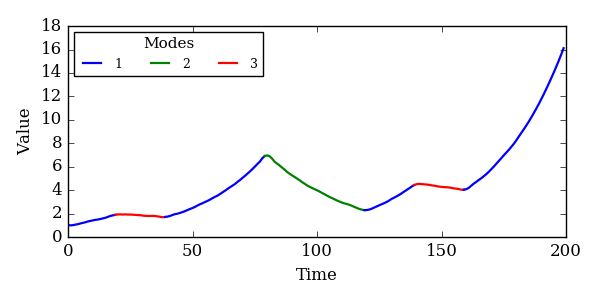}
\caption{\textbf{Trajectory observation} The trajectory that was fed into the segmentation algorithm. The colors represent the true underlying dynamical systems used to generate the trajectory parts. Note that only the first dimension of the state sequence is shown.}
\label{fig:real-traj}
\end{figure}
\begin{figure}
\centering
\includegraphics[width=\linewidth]{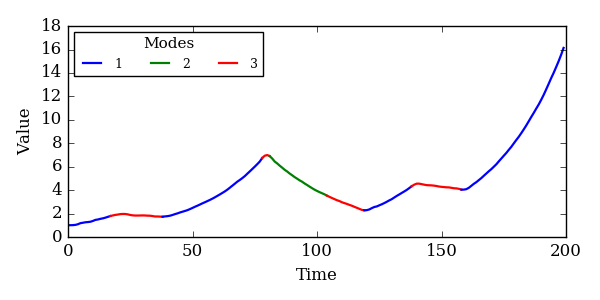}
\caption{\textbf{Sampled segmentation} The segmentation of the trajectory returned by the algorithm. Note that only the first dimension of the state sequence is shown.}
\label{fig:real-traj}
\end{figure}
The nonparametric part comes into play in the sense that we initialized the truncation level of the Dirichlet distribution used to approximate the DP with $L=100$. Still, the sampler correctly inferred that only three modes are needed to sufficiently describe the given dynamical system. The sampler was only able to recover distributions with an enforced self-transition probability due to the introduction of the sticky parameter $\kappa$. To show the impact of the sticky property, the algorithm was run on the same trajectory with the same setup, but this time with setting $\kappa = 0$. The result is shown in Figure \ref{fig:sampled-modes-nonsticky}.

It can be seen that the sampled mode sequence now displays high-frequency switching between the modes which is undesirable since the overarching goal is to specifically find transition points to yield a trajectory segmentation. Fast switching modes do not provide sufficient temporal continuity to offer any meaningful segmentation for a given trajectory.



\section{Conclusion}

In our project, we have implemented an algorithm for learning an SLDS model to infer a meaningful segmentation of trajectories. This segmentation can be used, e.g., as an initial segmentation for algorithms such as ProbS, obviating the need to rely on simple heuristic methods. We have followed a nonparametric approach to infer the number of modes in the system and utilized a Hierarchical Dirichlet Process \cite{teh2004sharing} to introduce shared dynamical parameters among all modes. Furthermore, we have used the sticky extension introduced by Fox\,et\,al.\ \cite{Ebfox2009slds} to avoid high-frequent switching behavior. As learning an SLDS model in closed form has shown to be intractable, we have employed a Gibbs sampler iteratively estimating the quantities of the model following \cite{Ebfox2009slds}. The modes $z_t$, states $x_t$, the hyperparameters and dynamical parameters $\theta = \{ \mat{\Sigma}^{(k)}, \mat{A}^{(k)}\}$ are alternately sampled to provide an estimate for the joint distribution of all variables for the SLDS after a burn-in phase.

The algorithm was tested in a toy example to provide a segmentation for a low-dimensional trajectory generated by an SLDS. The parameters of the system were tried to be recovered by applying the algorithm to a noisy observation of the generated trajectory. After a short burn-in phase, the algorithm was able to infer nearly every switching point correctly.

The results of the experiment show that the nonparametric SLDS model is a potential alternative for segmenting demonstration trajectories by finding appropriate mode transition points. However, tuning and optimization of the hyperparameters proved to be a rather tedious and unintuitive task. 
In the future, we want to synthesize the nonparametric segmentation approach with the ProbS algorithm by Lioutikov\,et\,al.\ \cite{lioutikov2015probabilistic} in order to boost the performance by providing meaningful initialization segmentations.


\bibliography{main}
\bibliographystyle{ieeetr}

\end{document}